\documentclass[10pt,twocolumn,letterpaper]{article}

\usepackage{cvpr}
\usepackage{times}
\usepackage{epsfig}
\usepackage{graphicx}
\usepackage{tabularx}
\usepackage{amsmath}
\usepackage{amssymb}
\usepackage[nocompress]{cite}

\usepackage[pagebackref=true,breaklinks=true,letterpaper=true,colorlinks,bookmarks=false]{hyperref}

\cvprfinalcopy % *** Uncomment this line for the final submission

\def\cvprPaperID{110} % *** Enter the CVPR Paper ID here

\DeclareMathOperator*{\argmin}{arg\,min}

\newcommand{\specialcell}[2][c]{\begin{tabular}[#1]{@{}c@{}}#2\end{tabular}}

\ifcvprfinal\fi
\begin{document}

%%%%%%%%% TITLE
\title{Learning Deep Feature Representations with Domain Guided Dropout for Person Re-identification}

\author{Tong Xiao \qquad Hongsheng Li \qquad Wanli Ouyang \qquad Xiaogang Wang\\
Department of Electronic Engineering, The Chinese University of Hong Kong\\
{\tt\small \{xiaotong,hsli,wlouyang,xgwang\}@ee.cuhk.edu.hk}}

\maketitle
\thispagestyle{empty}

%%%%%%%%% ABSTRACT
\begin{abstract}
Learning generic and robust feature representations with data from multiple domains for the same problem is of great value, especially for the problems that have multiple datasets but none of them are large enough to provide abundant data variations. In this work, we present a pipeline for learning deep feature representations from multiple domains with Convolutional Neural Networks (CNNs). When training a CNN with data from all the domains, some neurons learn representations shared across several domains, while some others are effective only for a specific one. Based on this important observation, we propose a Domain Guided Dropout algorithm to improve the feature learning procedure. Experiments show the effectiveness of our pipeline and the proposed algorithm. Our methods on the person re-identification problem outperform state-of-the-art methods on multiple datasets by large margins.
\end{abstract}

%%%%%%%%% BODY TEXT
\section{Introduction} % (fold)
\label{sec:introduction}

In computer vision, a \emph{domain} often refers to a dataset where samples follow the same underlying data distribution. It is common that multiple datasets with different data distributions are proposed to target the same or similar problems. Multi-domain learning aims to solve the problem with datasets across different domains simultaneously by using all the data they provide. As deep learning arises in the recent years, learning good feature representations achieves great success in many research fields and real-world applications. The success of deep learning is driven by the emergence of large-scale training data, which makes multi-domain learning an interesting problem. Many studies~\cite{donahue2013decaf,azizpour2014generic,oquab2014learning} have shown that fine-tuning a deep model pretrained on a large-scale dataset (\eg ImageNet~\cite{deng2009imagenet}) is effective for other related domains and tasks. However, in many specific areas, there is no such large-scale dataset for learning robust and generic feature representations. Nonetheless, different research groups have proposed many smaller datasets. It is necessary to develop an effective algorithm that jointly utilizes all of them to learn generic feature representations.

\begin{figure}[t]
\begin{center}
\includegraphics[width=1.0\linewidth]{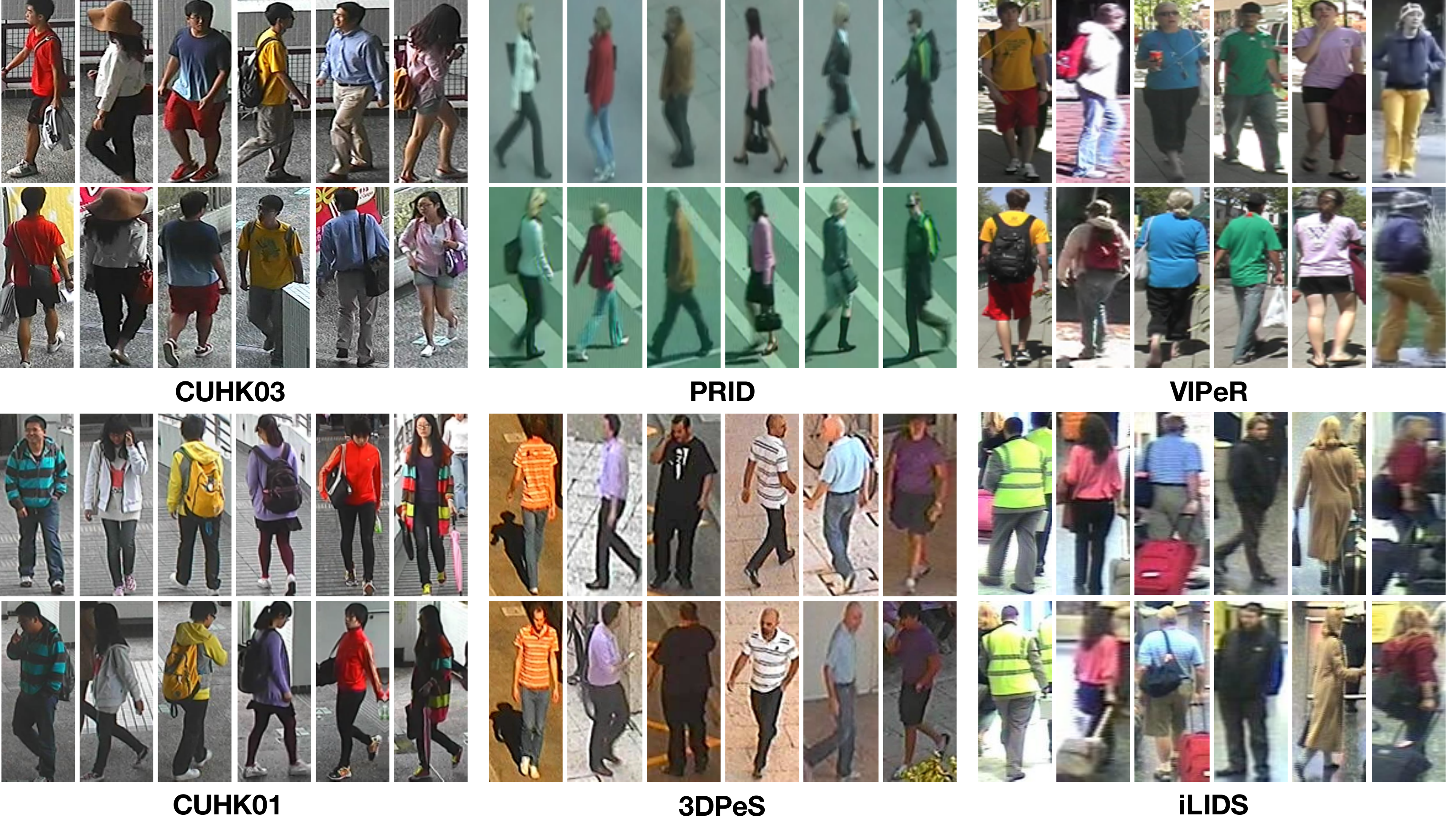}
\end{center}
\caption{Examples of multiple person re-identification datasets. Each dataset has its own bias. Our goal is to learn generic feature representations that are effective on all of them simultaneously.}
\label{fig:reid_datasets}
\end{figure}

Another interesting aspect of multi-domain learning is that it enriches the data variety because of the domain discrepancies. Limited by various conditions, data collected by a research group might only include certain types of variations. Take the person re-identification~\cite{ajakan2014domain,li2014deepreid} problem as an example, pedestrian images are usually captured in different scenes (\eg, campus, markets, and streets), as shown in Figure~\ref{fig:reid_datasets}. Images in CUHK01~\cite{li2013locally} and CUHK03~\cite{li2014deepreid} are captured on campus, where many students wear backpacks. PRID~\cite{hirzer11a} contains pedestrians in street views, where crosswalks appear frequently in the dataset. Images in VIPeR~\cite{gray2007evaluating} suffer from significant resolution changes across different camera views. Each of such datasets is biased and contains only a subset of possible data variations, which is not sufficient for learning generic feature representations. Combining them together can diversify the training data, thus makes the learned features more robust.

In this paper, we present a pipeline for learning generic feature representations from multiple domains that are effective on all of them simultaneously. For concrete demonstration, we target the person re-identification problem, but the method itself would be generalized to other problems with datasets of multiple domains. As learning features from a large-scale classification dataset is proved to be effective~\cite{sun2014deep}, we first mix all the domains together and train a Convolutional Neural Network (CNN) to recognize person identities (IDs). The CNN model we designed consists of several BN-Inception~\cite{szegedy2014going,ioffe2015batch} modules, and its capacity well fits to the scale of the mixed dataset. This carefully designed CNN model provides us a fairly strong baseline, but the simple joint learning scheme does not take full advantages of the variations of multiple domains.

Intuitively, neurons that are effective for one domain could be useless for another domain because of the presence of domain biases. For example, only the i-LIDS dataset contains pedestrians with luggages, thus the neurons that capture luggage features are of no use when recognizing people from other domains.

Based on this observation, we propose \emph{Domain Guided Dropout} --- a simple yet effective method of muting non-related neurons for each domain. Different from the standard Dropout~\cite{hinton2012improving}, which treats all the neurons equally, our method assigns each neuron a specific dropout rate for each domain according to its effectiveness on that domain. The proposed Domain Guided Dropout has two schemes, a deterministic scheme, and a stochastic scheme. After the baseline model is trained jointly with datasets of all the domains, we replace the standard Dropout with the deterministic Domain Guided Dropout and resume the training for several epochs. We observe that the proposed dropout scheme consistently improves the performance on all the domains after several epochs, especially on the smaller-scale ones. This step produces better generic feature representations that are effective on all the domains simultaneously. We further fine-tune the net with stochastic Domain Guided Dropout on each domain separately to obtain the best possible results.

The contribution of our work is three-fold. First, we present a pipeline for learning generic feature representations from multiple domains that perform well on all of them. This enables us to learn better features from multiple datasets for the same problem. Second, we propose Domain Guided Dropout to discard useless neurons for each domain, which improves the performance of the CNN. At last, our method outperforms state-of-the-arts on multiple person re-identification datasets by large margins. We observe that learning feature representations by utilizing data from multiple datasets improve the performance significantly, and the largest gain is 46\% on the PRID dataset. Extensive experiments validate our proposed method and the internal mechanism of the method is studied in details.

% section introduction (end)

\section{Related Work} % (fold)
\label{sec:related_work}

In recent years, training deep neural networks with multiple domains has been explored. Feature representations learned by Convolutional Neural Networks have shown their effectiveness in a wide range of visual recognition tasks~\cite{krizhevsky2012imagenet,girshick2014rich,long2014fully,chu2016structure,yang2016end,kang2016object}. Long~\etal~\cite{long2015learning} incorporated the multiple kernel variant of Maximum Mean Discrepancy (MMD) objective for regularizing the training of neural networks. Ganin~\etal~\cite{ganin2014unsupervised} proposed to reduce the distribution mismatch between the source and target domains by reversing the gradients of the domain classification loss, which is also utilized by~\cite{tzeng2015simultaneous} with a softlabel matching loss to transfer task information. Most of these methods aim at finding a common feature space that is domain invariant. However, our approach allows the representation to have disjoint components that are domain specific, while also learning a shared representation.

As deep neural networks usually contain millions of parameters, it is of great importance to reduce the parameter space by adding regularizations to the weights. The quality of the regularization method would significantly affect both the discriminative power and generalization ability of the trained networks. Dropout~\cite{hinton2012improving} is one of the most widely used regularization method in training deep neural networks, which significantly improves the performance of the deep model~\cite{krizhevsky2012imagenet}. During the network training process, Dropout randomly sets neuron responses to zero with a probability of 0.5. Thus a training batch updates only a subset of all the neurons at each time, which avoids co-adaptation of the learned feature representations.

While the standard Dropout algorithm treats all the neurons equally with a fixed probability, Ba~\etal~\cite{ba2013adaptive} proposed an adaptive dropout scheme by learning a binary belief network to predict the dropout probability for each neuron. In practice, they use the response of each neuron to compute the dropout probability for itself. Our approach significantly differs from this method, as we propose to train a CNN from multiple domains, and utilize the domain information to guide the dropout procedure.

We target the person re-identification (Re-ID) problem in this work, which is very challenging and draws much attention in recent years~\cite{wang2007shape,li2015locality,zhang2015bit,xu2013human,liu2014fly,li2015multi}. Existing Re-ID methods mainly address the problem from two aspects: finding more powerful feature representations and learning better metrics. Zhao~\etal~\cite{zhao2013unsupervised,zhao2013person,zhao2014learning} proposed to combine SIFT features with color histogram as features. In deep learning literature, Li~\etal~\cite{li2014deepreid} and Ahmed~\etal~\cite{ahmed2015improved} designed CNN models specifically to the Re-ID task and achieved good performance on large-scale datasets. They trained the network with pairs of pedestrian images and adopted the verification loss function. Ding~\etal~\cite{ding2015deep} utilized triplet samples for training features that maximize relative distance between the pair of same person and the pair of different people in the triplets. Apart from the feature learning methods, a large number of metric learning algorithms~\cite{paisitkriangkrai2015learning,weinberger2005distance,xiong2014person,davis2007information,mcfee2010metric,koestinger2012large} have also been proposed to solve the Re-ID problem from a complementary perspective. Some recent works addressed the problem of mismatch between traditional Re-ID and real application scenarios. Liao~\etal~\cite{liao2014open} proposed a database for open-set Re-ID. Zheng~\etal~\cite{zheng2015person} treated Re-ID as an image search problem and introduced a large-scale dataset. Xu~\etal~\cite{xu2014person} raised the problem of searching a person inside whole images rather than cropped bounding boxes.

% section related_work (end)

\section{Method} % (fold)
\label{sec:method}

\begin{figure*}[t]
\begin{center}
\includegraphics[width=0.9\linewidth]{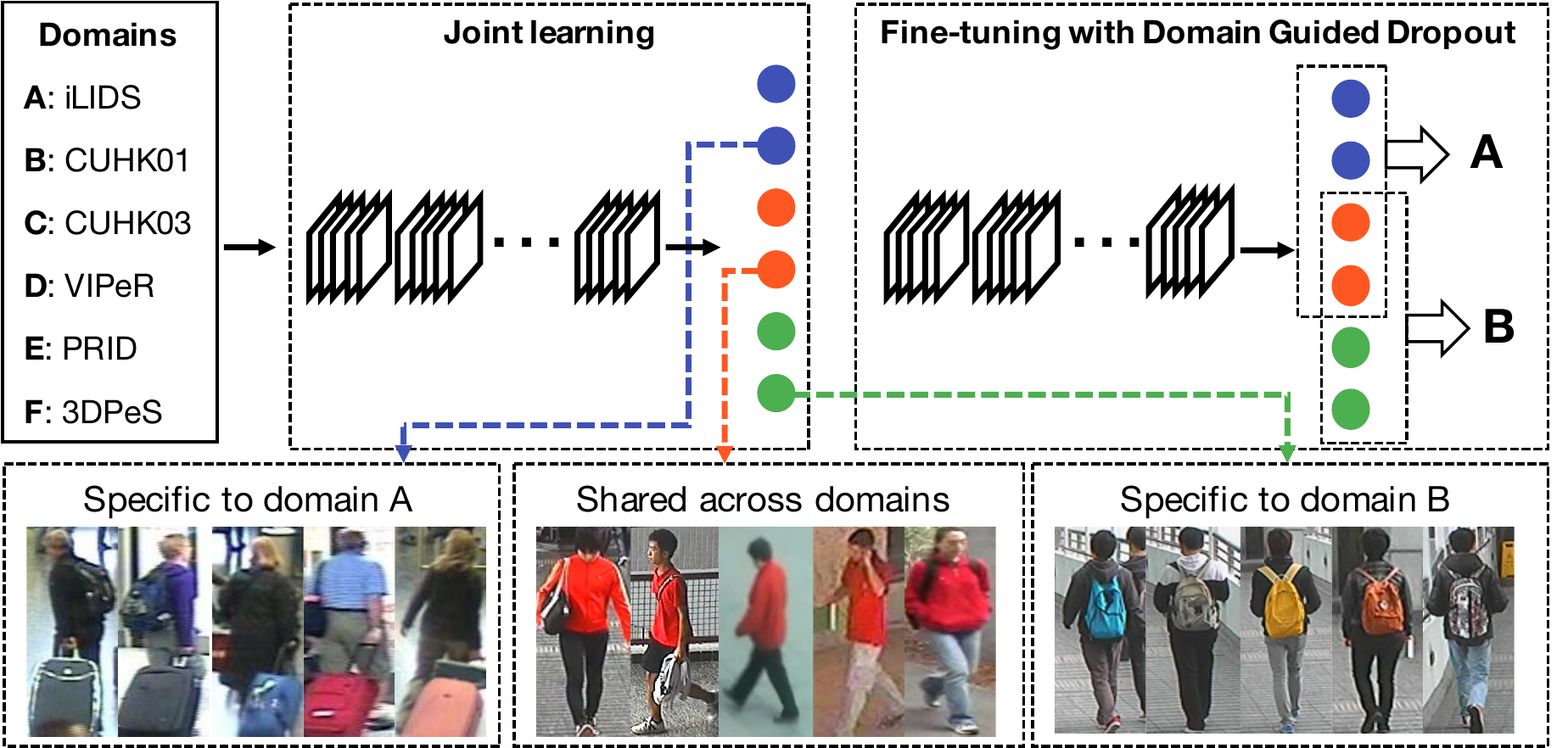}
\end{center}
\caption{Overview of our pipeline. For the person re-identification problem, we first train a CNN jointly on all six domains. Then we analyze the effectiveness of each neuron on each domain. For example, some may capture the luggages that only appear in domain $A$, while some others may capture the red clothes shared across different domains. We propose a Domain Guided Dropout algorithm to discard useless neurons for each domain during the training process, which drives the CNN to learn better feature representations on all the domains simultaneously.}
\label{fig:overview}
\end{figure*}

Our proposed pipeline for learning CNN features from multiple domains consists of several stages. As shown in Figure~\ref{fig:overview}, we first mix the data and labels from all the domains together, and train a carefully designed CNN from scratch on the joint dataset with a single softmax loss. This pretraining step produces a strong baseline model that works on all the domains simultaneously. Next, for each domain, we perform the forward pass on all its samples and compute for each neuron its average impact on the objective function. Then we replace the standard Dropout layer with the proposed Domain Guided Dropout layer, and continue to train the CNN model for several more epochs. With the guidance of which neurons being effective for each domain, the CNN learns more discriminative features for all of them. At last, if we want to obtain feature representations for a specific domain, the CNN could be further fine-tuned on it, again with the Domain Guided Dropout to improve the performance. In this section, we detail these stages, and compare our design choices with other alternatives.

\subsection{Problem formulation} % (fold)
\label{sub:problem_formulation}

Although the pipeline itself is not limited to any specific scope, we target the person re-identification problem for concrete demonstration. The problem can be formulated as follows. Suppose we have $D$ domains, each of which consists of $N_i$ images of $M_i$ different people. Let $\{(x_i^{(j)}, y_i^{(j)})_{j=1}^{N_i}\}_{i=1}^{D}$ denote all training samples, where $x_i^{(j)}$ is the $j$-th image of the $i$-th domain, and $y_i^{(j)} \in \{1,2,\dots,M_i\}$ is the identity of the corresponding person. Our goal is to learn a generic feature extractor $g(\cdot)$ that has similar outputs for images of the same person and dissimilar outputs for different people. During the test phase, given a probe pedestrian image and a set of gallery images, we use $g(\cdot)$ to extract features from all of them, and rank the gallery images according to their Euclidean distances to the probe image in the feature space. For the training phase, there are several frameworks that use pairwise~\cite{li2014deepreid,ahmed2015improved} or triplet~\cite{schroff2015facenet} inputs for learning feature embeddings. In our approach, we train a CNN to recognize the identity of each person, which is also adopted in the face verification work~\cite{sun2014deep}.

% subsection problem_formulation (end)

\subsection{Joint learning objective and the CNN structure} % (fold)
\label{sub:joint_learning_and_cnn_structure}

When mixing all the $D$ domains together, a straightforward solution is to employ a multi-task objective function, \ie, learning $D$ softmax classifiers $f_1, f_2, \dots, f_D$ and a shared features extractor $g$ that minimize
\begin{equation} \label{eq:multi-task-objfunc}
    \argmin_{f_1,f_2,\dots,f_D,g} \sum_{i=1}^D \sum_{j=1}^{N_i} \mathcal{L}\left( f_i(g(x_i^{(j)})), y_i^{(j)} \right),
\end{equation}
where $\mathcal{L}$ is the softmax loss function that equals to the cross-entropy between the predicted probability vector and the ground truth.

However, since different person re-identification datasets usually have totally different identities, it is also safe to merge all $M=\sum_{i=1}^D M_i$ people together and relabel them with new IDs ${y'} \in \{1,2,\dots,M\}$. For the merged dataset, we can define a single-task objective function, \ie, learning one softmax classifier $f$ and the features extractor $g$ that minimize
\begin{equation} \label{eq:single-task-objfunc}
    \argmin_{f,g} \sum_{i=1}^D \sum_{j=1}^{N_i} \mathcal{L}\left( (f\circ g)(x_i^{(j)}), {y'}_i^{(j)} \right).
\end{equation}

Compared with the multi-task formulation, this single-task learning scheme forces the network to simultaneously distinguish people from all domains. The feature representations capture two types of information: domain biases (\eg, background clutter, lighting, \etc) as well as person appearance and attributes. If the data distributions of two domains differ a lot, it would be easy to separate the persons of the two domains by observing only the domain biases. However, when these biases are not significant enough, the network is required to learn discriminative person-related features to make the decisions. Thus the single-task objective fits better to our setting and is chosen for this work.

Since pedestrian images are usually quite small and are not of square-shapes, it is not appropriate to directly use the ImageNet pretrained CNN models, which are trained with object images of high resolution and abundant details. Thus we propose to design a network structure that well fits our problem scale. Inspired by~\cite{ioffe2015batch,simonyan2014very}, we build a CNN with three preceding $3\times 3$ convolutional layers followed by six Inception modules and two fully connected layers. Detailed structures are listed in Table~\ref{tab:cnn_structure}. The Batch Normalization (BN) layers are employed before each ReLU layer, which accelerate the convergence process and avoid manually tweaking the initialization of weights and biases. For training the CNN from scratch, we randomly dropout 50\% neurons of the fc7 layer. The initial learning rate is set to 0.1 and is decreased by 4\% for every 4 epochs until it reaches 0.0005. The learning rate is then fixed at this value for a few more epochs until convergence.

\begin{table*}[t]
\begin{center}
\begin{tabular}{|l|c|c|c|c|c|c|c|c|}
\hline
name & \specialcell{patch size/\\stride} & \specialcell{output\\size} & \#1$\times$1 & \specialcell{\#3$\times$3\\reduce} & \#3$\times$3 & \specialcell{double \#3$\times$3\\reduce} & \specialcell{double\\\#3$\times$3} & pool+proj \\
\hline\hline
input & & $3\times 144\times 56$ & & & & & & \\
\hline
conv1 -- conv3 & $3\times 3/2$ & $32\times 144\times 56$ & & & & & & \\
% \hline
% conv2 & $3\times 3/2$ & $32\times 144\times 56$ & & & & & & \\
% \hline
% conv3 & $3\times 3/2$ & $32\times 144\times 56$ & & & & & & \\
\hline
pool3 & $2\times 2/2$ & $32\times 72\times 28$ & & & & & & \\
\hline
inception (4a) & & $256\times 72\times 28$ & 32 & 32 & 32 & 32 & 32 & avg + 32 \\
\hline
inception (4b) & stride 2 & $384\times 72\times 28$ & 32 & 32 & 32 & 32 & 32 & max + pass through \\
\hline
inception (5a) & & $512\times 36\times 14$ & 64 & 64 & 64 & 64 & 64 & avg + 64 \\
\hline
inception (5b) & stride 2 & $768\times 36\times 14$ & 64 & 64 & 64 & 64 & 64 & max + pass through \\
\hline
inception (6a) & & $1024\times 36\times 14$ & 128 & 128 & 128 & 128 & 128 & avg + 128 \\
\hline
inception (6b) & stride 2 & $1536\times 36\times 14$ & 128 & 128 & 128 & 128 & 128 & max + pass through \\
\hline
fc7 & & $256$ & & & & & & \\
\hline
fc8 & & $M$ & & & & & & \\
\hline\noalign{\smallskip}
\end{tabular}
\end{center}
\caption{The structure of our proposed CNN for person re-identification}
\label{tab:cnn_structure}
\end{table*}

% subsection joint_learning_and_cnn_structure (end)

\subsection{Domain Guided Dropout} % (fold)
\label{sub:guided_dropout}

\begin{figure}[t]
\begin{center}
\includegraphics[width=1.0\linewidth]{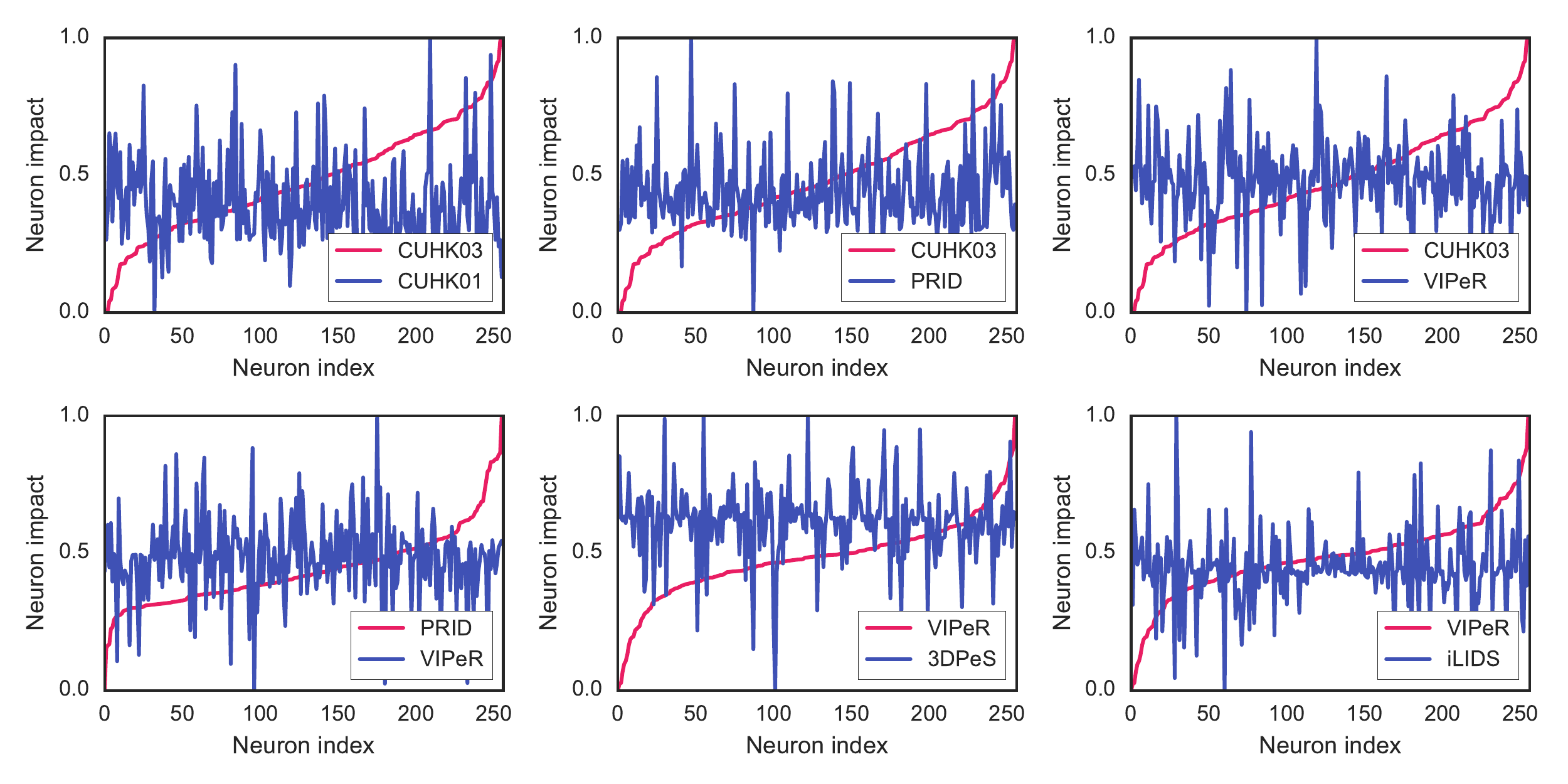}
\end{center}
\caption{The neuron impact scores between several pairs of domains. For each pair of domains $(A,B)$, the neurons are sorted w.r.t. their impact scores on domain $A$ (red curves). Their impact scores on domain $B$ are shown in blue. The two curves have little correlation, which indicates that different domains have different effective neurons.}
\label{fig:impact_between_domains}
\end{figure}

Given the CNN model pretrained by using the mixed dataset, we identify for each domain which neurons are effective. For each domain sample, we define the impact of a particular neuron on this sample as the gain of the loss function when we remove the neuron. Specifically, let $g(x)\in \mathbb{R}^d$ denote the $d$-dimensional CNN feature vector of an image $x$. The impact score of the $i$-th $(i\in\{1,2,\dots,d\})$ neuron on this image sample is defined as
\begin{equation} \label{eq:true_scores}
   s_i=\mathcal{L}(g(x)_{\setminus i}) - \mathcal{L}(g(x)),
\end{equation}
where $g(x)_{\setminus i}$ is the feature vector after we setting the $i\text{-th}$ neuron response to zero. For each domain $\mathcal{D}$, we then take the expectation of $s_i$ over all its samples to obtain the averaged impact score $\bar{s}_i=\mathrm{E}_{x\in\mathcal{D}}[s_i]$. We visualize the neuron impact scores between several pairs of domains in Figure~\ref{fig:impact_between_domains}. It clearly shows that the two sets of impact scores have little correlation, indicating that the effective neurons for different domains are not the same.

\begin{figure}[t]
\begin{center}
\includegraphics[width=1.0\linewidth]{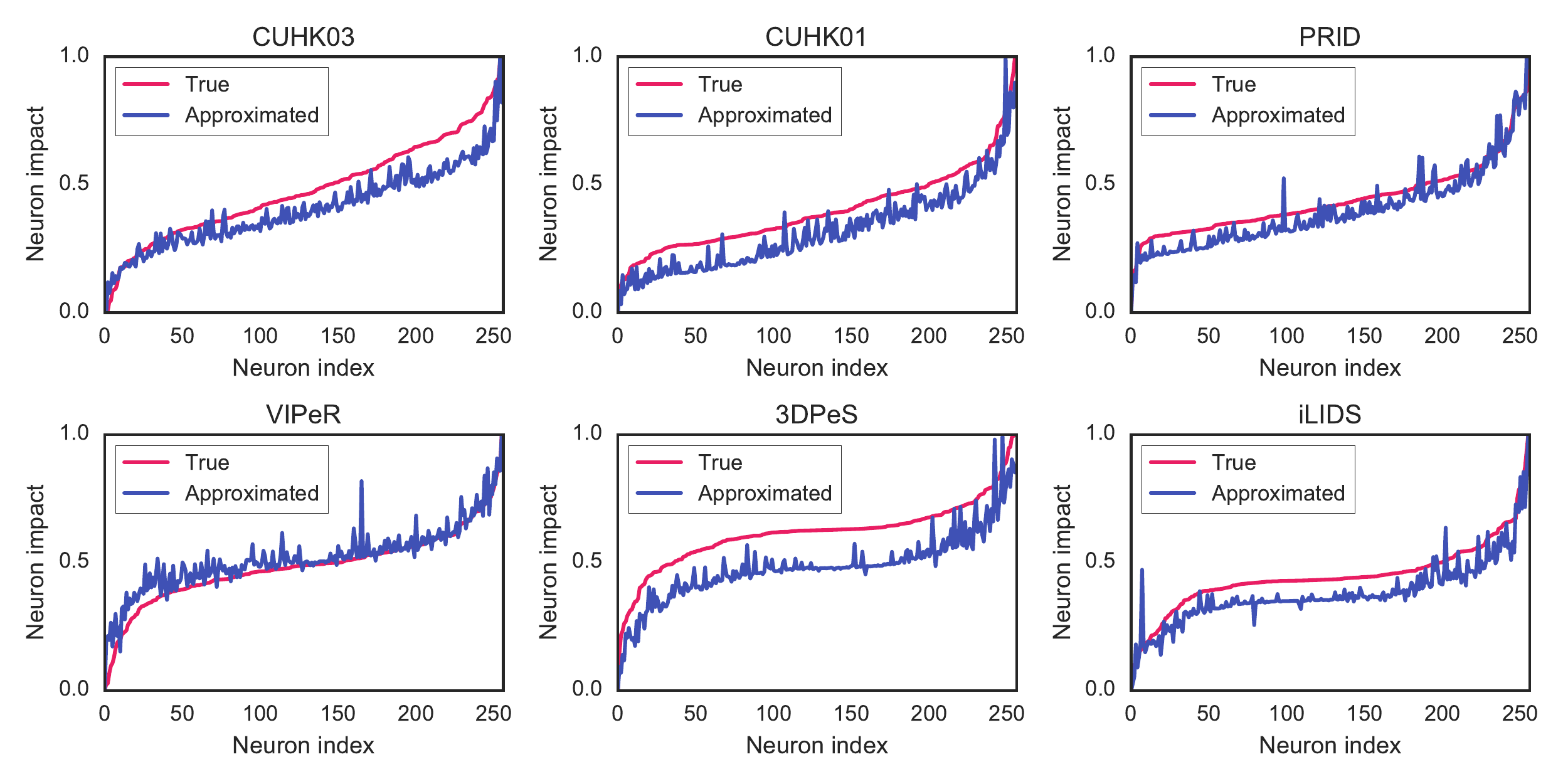}
\end{center}
\caption{Comparison of the true (Eq.~\eqref{eq:true_scores}) and the approximated (Eq.~\eqref{eq:approx_scores}) neuron impact scores}
\label{fig:impact_true_vs_approx}
\end{figure}

A naive computation of all the impact values requires $O(d|\mathcal{D}|)$ network forward passes, which is quite expensive if $d$ is large. Therefore, we follow~\cite{simonyan2013deep} to accelerate the process by using approximate Taylor's expansion of $\mathcal{L}(g(x))$ to the second order
\begin{equation} \label{eq:approx_scores}
   s_i\approx -\frac{\partial \mathcal{L}}{\partial g(x)_i} g(x)_i + \frac{1}{2} \frac{\partial^2 \mathcal{L}}{\partial g(x)_i^2} g(x)_i^2.
\end{equation}
We study the quality of this approximation empirically, and observe that it is more accurate for higher-level layers close to the loss function. Here we show in Figure~\ref{fig:impact_true_vs_approx} the difference between the approximation and its true values for the neurons of the fc7 layer.

After obtaining all the $\bar{s}_i$, we continue to train the CNN model, but with these impact scores as guidance to dropout different neurons for different domains during the training process. For all the samples belonging to a particular domain, we generate a binary mask $m$ for the neurons according to their impact scores $s$, and then elementwisely multiply $m$ with the neuron responses. Two schemes are proposed on how to generate the mask $m$. The first one is deterministic, which discards all the neurons having non-positive impact scores
\begin{equation}
   m_i=\begin{cases}
      1 & \quad \text{if } s_i > 0\\
      0 & \quad \text{if } s_i \le 0\\
   \end{cases}
\end{equation}

The other one is stochastic, where $m_i$ is drawn from a Bernoulli distribution with probability
\begin{equation} \label{eq:stochastic_guided_dropout}
   p(m_i=1) = \frac{1}{1+e^{-s_i/T}}.
\end{equation}
Here we use the sigmoid function to map a impact score to $(0,1)$, and $T$ is the temperature that controls how significantly the scores $s$ would affect the probabilities. When $T\to 0$, it is equivalent to the deterministic scheme; when $T\to \infty$, it falls back to the standard Dropout with a ratio of $0.5$. We study the effect of $T$ empirically in Section~\ref{sub:effectiveness_of_the_guided_dropout_module}.

We apply the Domain Guided Dropout to the fc7 neurons and resume the training process. The network's learning rate policy is changed to decay polynomially from 0.01 with the power parameter set to 0.5. The whole network is trained for 10 more epochs.

During the test stage, for the deterministic scheme, the neurons are also discarded if their impacts are no greater than zero. While for the stochastic scheme, we keep all the neuron responses but scale the $i$-th one with $1/(1+e^{-s_i/T})$.

% subsection guided_dropout (end)

% section method (end)

\section{Experiments} % (fold)
\label{sec:experiments}

We conducted experiments on several popular person re-identification datasets. In this section, we first detail the characteristics of each dataset and the test protocols we followed in Section~\ref{sub:datasets_and_protocols}. Then we compare the results of our approach with state-of-the-arts, showing the effectiveness of our multi-domain deep learning pipeline in Section~\ref{sub:comparison_with_state_of_the_art_results}. Section~\ref{sub:effectiveness_of_the_guided_dropout_module} analyzes the Domain Guided Dropout module through a series of experiments, and discusses its properties based on the results. At last, we present some figures that help us understand the underlying mechanisms. The code is publicly available on GitHub\footnote{\url{https://github.com/Cysu/person_reid}}.

\subsection{Datasets and protocols} % (fold)
\label{sub:datasets_and_protocols}

There exist many challenging person re-identification datasets. In our experiments, we chose seven of them to cover a wide range of domain varieties. CUHK03~\cite{li2014deepreid} is one of the most largest published person re-identification datasets, it consists of five different pairs of camera views, and has more than 14,000 images of 1467 pedestrians. CUHK01~\cite{li2013locally} is also captured on the same campus with CUHK03, but only has two camera views and 1552 images in total. PRID~\cite{hirzer11a} extracts pedestrian images from recorded trajectory video frames. It has two camera views, each contains 385 and 749 identities, respectively. But only 200 of them appear in both views. Shinpuhkan~\cite{kawanishi2014shinpuhkan2014} is another large-scale dataset with more than 22,000 images. The highlight of this dataset is that it contains only 24 individuals, but all of them are captured with 16 cameras, which provides rich information on intra-personal variations.

The remaining three datasets are relatively quite small. VIPeR~\cite{gray2007evaluating} is one of the most challenging dataset, since it has 632 people but with various poses, viewpoints, image resolutions, and lighting conditions. 3DPeS~\cite{baltieri20113dpes} has 193 identities but the number of images for each person is not fixed. iLIDS~\cite{zheng2009associating} captures 119 individuals by surveillance cameras in an airport, and thus consists of large occlusions due to luggages and other passengers.

Since Shinpuhkan dataset has only 24 people, it cannot be used for testing the performance of re-identification systems. Thus we only use it in the training phase. For the other datasets, we mainly follow the settings in~\cite{paisitkriangkrai2015learning} to generate the test probe and gallery sets. But our training set has two differences with theirs. First, both the manually cropped and automatically detected images in CUHK03 were used. Second, we sampled 10 images from the video frames of the training identities in PRID. We also randomly drew roughly 20\% of all these images for validation. Notice that both the training and validation identities have no overlap with the test ones. The statistics of all the datasets and evaluation protocols are summarized in Table~\ref{tab:datasets}. In our experiments, we employed the commonly used CMC~\cite{moon2001computational} top-1 accuracy to evaluate all the methods.

\begin{table}[t]
\small
\begin{center}
\begin{tabular}{lccccc}
\hline\noalign{\smallskip}
Dataset & \specialcell{\#ID} & \specialcell{\#Trn.\\images} & \specialcell{\#Val.\\images} & \specialcell{\#Prb.\\ID} & \specialcell{\#Gal.\\ID} \\
\noalign{\smallskip}\hline\hline\noalign{\smallskip}
CUHK03~\cite{li2014deepreid} & 1467 & 21012 & 5252 & 100 & 100 \\
CUHK01~\cite{li2013locally} & 971 & 1552 & 388 & 485 & 485 \\
PRID~\cite{hirzer11a} & 385 & 2997 & 749 & 100 & 649 \\
VIPeR~\cite{gray2007evaluating} & 632 & 506 & 126 & 316 & 316\\
3DPeS~\cite{baltieri20113dpes} & 193 & 420 & 104 & 96 & 96 \\
i-LIDS~\cite{zheng2009associating} & 119 & 194 & 48 & 60 & 60 \\
Shinpuhkan~\cite{kawanishi2014shinpuhkan2014} & 24 & 18004 & 4500 & & \\
\hline\noalign{\smallskip}
\end{tabular}
\end{center}
\caption{Statistics of the datasets and evaluation protocols}
\label{tab:datasets}
\end{table}

% subsection datasets_and_protocols (end)

\subsection{Comparison with state-of-the-art methods} % (fold)
\label{sub:comparison_with_state_of_the_art_results}

We compare the results of our approach with those by state-of-the-art ones on all the six test datasets. For the 3DPeS and iLIDS datasets, the best previous method are~\cite{xiong2014person} and~\cite{ding2015deep}, respectively. While for the other four datasets, the best results are reported by~\cite{paisitkriangkrai2015learning}. Both methods are built upon hand-crafted features, and exploit a ranking ensemble of kernel-based metrics to boost the performance. However, our method relies on the learned CNN features and uses the Euclidean distance directly as the metric, which stresses the quality of the learned features representation rather than the metrics.

In order to validate our approach, we first obtain a baseline by training the CNN individually on each domain. Then we merge all the domains jointly with a single-task learning objective (JSTL) and train the CNN from scratch using all these domains. Next, we improve the learned CNN with the proposed deterministic Domain Guided Dropout (JSTL+DGD). Notice that this step provides a single model working on all the domains simultaneously. To show our best possible results, we further fine-tune the CNN separately on each domain with the stochastic Domain Guided Dropout (FT-JSTL+DGD). We also adopt a baseline method by fine-tuning from the JSTL model on each domain with standard dropout (FT-JSTL) for comparison. The results are summarized in Table~\ref{tab:comparison_with_sota}.

\begin{table}[t]
\small
\begin{center}
\begin{tabular}{llll}
\hline\noalign{\smallskip}
\noalign{\smallskip}
Method & CUHK03 & CUHK01 & PRID\\
\noalign{\smallskip}\hline\hline\noalign{\smallskip}
Best & 62.1~\cite{paisitkriangkrai2015learning} & 53.4~\cite{paisitkriangkrai2015learning} & 17.9~\cite{paisitkriangkrai2015learning} \\
Individually & 72.6 & 34.4 & 37.0 \\
JSTL & 72.0 & 62.1 & 59.0 \\
JSTL+DGD & 72.5 & 63.0 & 60.0 \\
FT-JSTL & 74.8 & 66.2 & 57.0 \\
FT-JSTL+DGD & \textbf{75.3} & \textbf{66.6} & \textbf{64.0} \\
\hline\noalign{\smallskip}
\noalign{\smallskip}
Method & VIPeR & 3DPeS & iLIDS \\
\noalign{\smallskip}\hline\hline\noalign{\smallskip}
Best & \textbf{45.9}~\cite{paisitkriangkrai2015learning} & 54.2~\cite{xiong2014person} & 52.1~\cite{ding2015deep} \\
Individually & 12.3 & 31.1 & 27.5 \\
JSTL & 35.4 & 44.5 & 56.9 \\
JSTL+DGD & 37.7 & 45.6 & 59.6 \\
FT-JSTL & 37.7 & 54.0 & 61.1 \\
FT-JSTL+DGD & 38.6 & \textbf{56.0} & \textbf{64.6} \\
\hline\noalign{\smallskip}
\end{tabular}
\end{center}
\caption{CMC top-1 accuracies of different methods}
\label{tab:comparison_with_sota}
\end{table}

\begin{figure}[t]
\begin{center}
\includegraphics[width=0.8\linewidth]{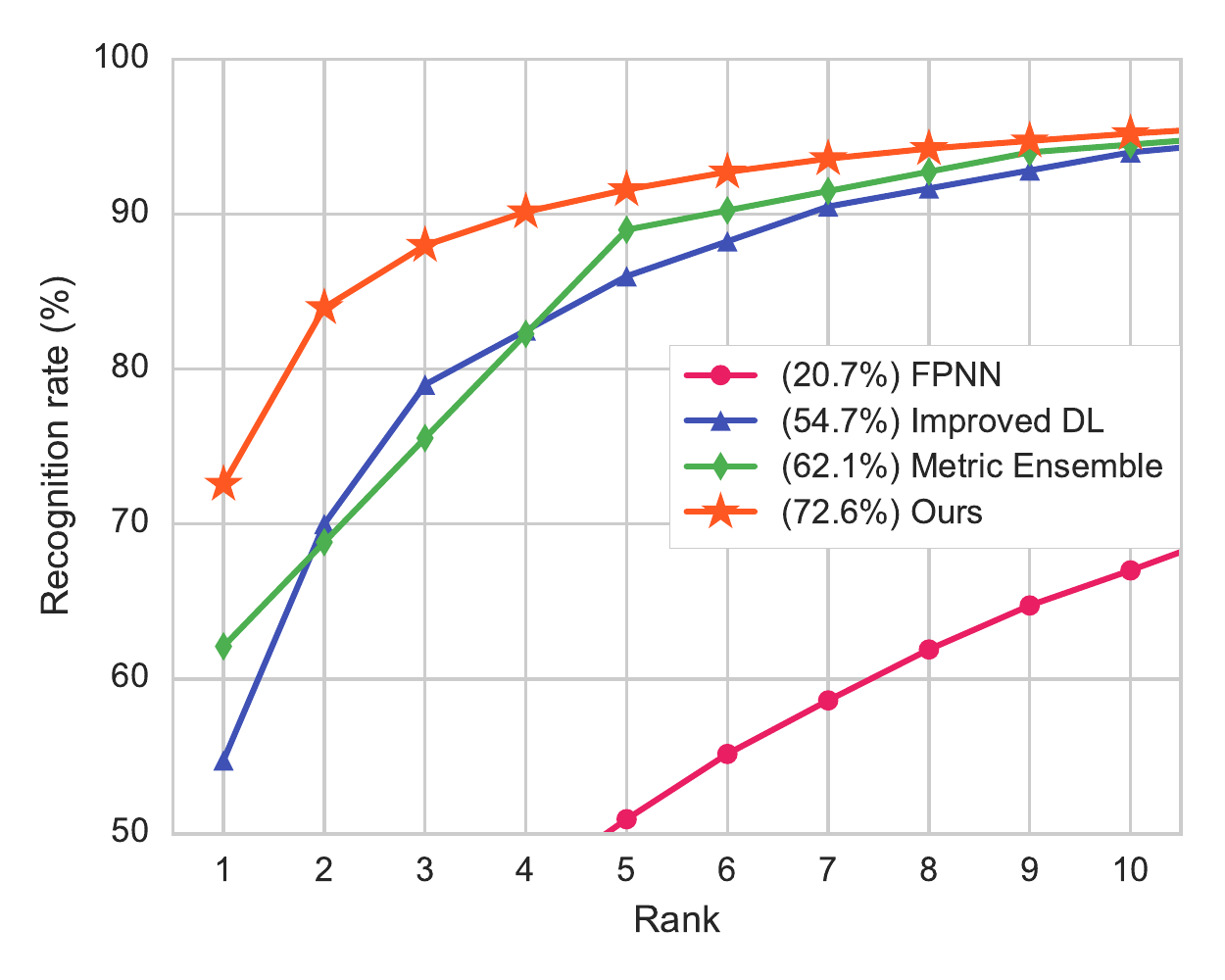}
\end{center}
\caption{CMC curves of different methods on CUHK03 dataset}
\label{fig:cmc_cuhk03}
\end{figure}

\textbf{CNN structure.} We first evaluate the effectiveness of the proposed CNN structure. When the network is trained only with the CUHK03 dataset, which is large enough for training CNN from scratch, we improve the state-of-the-art result by more than 10\% to 72.6\% (row 2 of Table~\ref{tab:comparison_with_sota}). Compared with the previous best deep learning method~\cite{ahmed2015improved}, whose result is 54.7\%, our method achieves a gain of 18\% in the performance. A two-stream network is used in~\cite{ahmed2015improved} to compute the verification loss given a pair of images, while we opt for learning a single CNN through an ID classification task and directly computing Euclidean distance based on the features. When the training set is large enough, this classification objective makes the CNN much easier to train. The CMC curves of different methods on the CUHK03 dataset are shown in Figure~\ref{fig:cmc_cuhk03}. However, when the dataset is quite small, it would be insufficient to learn such a large capacity network from scratch, which is demonstrated in Table~\ref{tab:comparison_with_sota} by the results of training the CNN only on each of the VIPeR, 3DPeS, and iLIDS datasets.

\textbf{Joint learning.} To overcome the scale issue of small datasets, we propose to merge all the datasets jointly as a single-task learning (JSTL) problem. In row three of Table~\ref{tab:comparison_with_sota}, we can see the performance increase on most of the datasets. This indicates learning from multiple domains jointly is very effective to produce generic feature representations for all the domains. An interesting phenomenon is that the performance on CUHK03 decreases slightly. We hypothesize that when combining different datasets together without special treatment, the larger domains would leverage their information to help the learning on the others, which makes the features more robust on different datasets but less discriminative on the larger ones themselves. Note that we do not balance the data from multiple sources in a mini-batch, as it would give more weights on smaller datasets, which leads to severe overfitting.

\textbf{Domain Guided Dropout.} The fourth row of Table~\ref{tab:comparison_with_sota} shows the effectiveness of applying the proposed Domain Guided Dropout (DGD) to the JSTL scheme. Based on the JSTL pretrained model, we compute the neuron impact scores of the fc7 layer on different domains, replace the standard Dropout layer with the proposed deterministic Domain Guided Dropout layer, and continue to train the network for several epochs. Although the original JSTL model has already converged to a local minimum, utilizing Domain Guided Dropout consistently improves the performance on all the domains by 0.5\%-2.7\%. This indicates that it is effective to regularize the network specifically for different domains, which maximizes the discriminative power of the CNN on all the domains simultaneously.

At last, to achieve the best possible performance of our model on each domain, we fine-tune the previous JSTL+DGD model on each of them individually with stochastic Domain Guided Dropout. This step adapts the CNN to the specific domain biases and sacrifices the generalization ability to other domains. As a result, the final CMC top-1 accuracies are increased by several percents, as listed in the last row of Table~\ref{tab:comparison_with_sota}. On the other hand, comparing with FT+JSTL, the results are improved by 3\% on average, which indicates that JSTL+DGD provides better generic features. Note that FT+JSTL on PRID results in even worse performance than JSTL. Such overfitting problem is resolved by applying DGD.

\subsection{Effectiveness of Domain Guided Dropout} % (fold)
\label{sub:effectiveness_of_the_guided_dropout_module}

After evaluating the overall performance of our pipeline, we also investigate in details the effects of the proposed Domain Guided Dropout module in this subsection.

\begin{figure}[t]
\begin{center}
\includegraphics[width=0.9\linewidth]{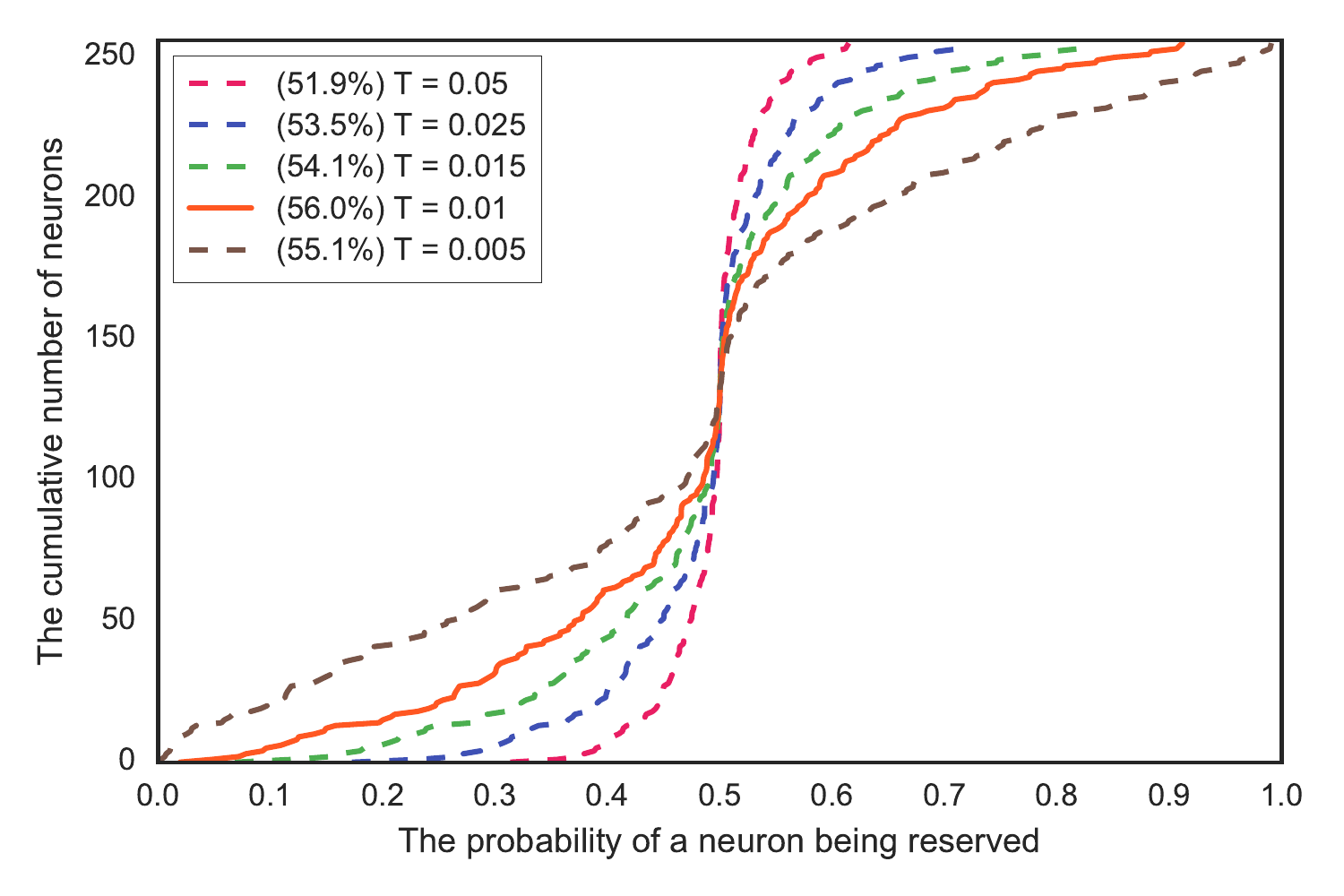}
\end{center}
\caption{The cumulative number of neurons to be reserved under certain probabilities. Different temperature $T$ settings and corresponding CMC top-1 accuracies are shown in the legend.}
\label{fig:temp_cdf}
\end{figure}

\textbf{Temperature $T$.} As the temperature $T$ significantly affects the behavior and performance of the stochastic Domain Guided Dropout scheme, we first study the effects of this hyperparameter. From the theoretical analysis we know that the stochastic Domain Guided Dropout falls back to the standard Dropout (ratio equals to 0.5) when $T\to \infty$, and to the deterministic scheme when $T\to 0$. However, it is still unclear how to set it properly in real applications. Therefore, we provide some empirical results of tuning the temperature $T$. We use the 3DPeS dataset as an example, and fine-tune the JSTL+DGD model on it with different values of $T$. For each temperature, all the fc7 neurons have certain probabilities to be reserved according to Eq~\eqref{eq:stochastic_guided_dropout}. We count the histogram of the neurons with respect to their probabilities to be reserved, and plot the cumulative distribution function in Figure~\ref{fig:temp_cdf}. We can see that the best performance can be achieved when $T$ is in a certain range that makes $\max_i p(m_i=1) \approx 0.9$. This phenomenon indicates that a good $T$ should assign the most effective neuron a high enough probability (0.9) to be reserved. We set $T$ according to this empirical observation when using the stochastic Domain Guided Dropout scheme  in our experiments.

\begin{figure}[t]
\begin{center}
\includegraphics[width=1.0\linewidth]{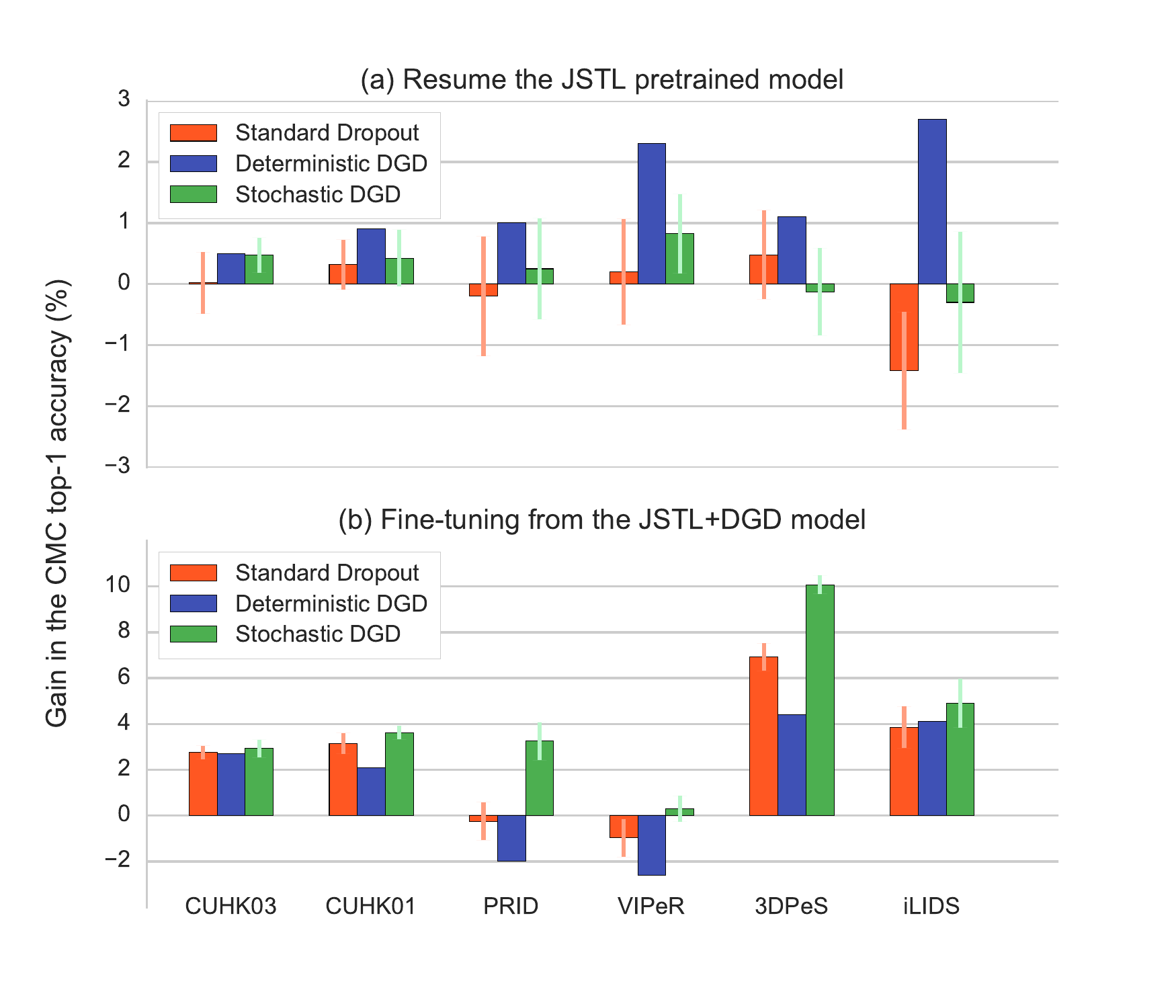}
\end{center}
\caption{Comparison of different Dropout schemes}
\label{fig:comparison_with_dropout}
\end{figure}

\textbf{Deterministic vs. stochastic.} The next question is whether the deterministic and stochastic Domain Guided Dropout have similar behaviors, or one outperforms the other in certain pipeline stages. We compare these two strategies within the JSTL+DGD and FT-JSTL+DGD stages in our pipeline. Their gains on the CMC performance for each domain under different settings are shown in Figure~\ref{fig:comparison_with_dropout} as the blue and green bars, respectively.

From Figure~\ref{fig:comparison_with_dropout}(a) we can see that when feeding the network with the data from all the domains, deterministic Domain Guided Dropout is better in general. This is because the objective here is to learn generic representations that are robust for different domains. The deterministic scheme strictly constrains that data from each domain are used to update only a specific subset of neurons. Thus it eliminates the potential confusion due to the discrepancies between different domains. On the contrary, when fine-tuning the CNN with the data only from one specific domain, the domain discrepancy no longer exists. All the inputs follow the same underlying distribution, so we can use stochastic Domain Guided Dropout to update all the neurons with proper guidance to determine the dropout rate for each of them, as shown in Figure~\ref{fig:comparison_with_dropout}(b). As a conclusion, the deterministic DGD is more effective when it is used to train the CNN jointly with all the domains, while the stochastic DGD is superior when fine-tuning the net separately on each domain.

\textbf{Standard Dropout vs. Domain Guided Dropout.} At last, we compare the proposed Domain Guided Dropout with the standard Dropout under different scenarios. The results are summarized in Figure~\ref{fig:comparison_with_dropout}. First, when resuming the training of the JSTL pretrained model, we applied the deterministic Domain Guided Dropout. From Figure~\ref{fig:comparison_with_dropout}(a) we can see that since the model is already converged, continue to use standard Dropout scheme cannot further improve the performance. The performance would rather jitter insignificantly or decrease on particular domains due to overfitting. However, by using the deterministic Domain Guided Dropout scheme, the performance improves consistently on all the domains, especially for the small-scale ones. On the other hand, by comparing the orange and the green bars in Figure~\ref{fig:comparison_with_dropout}(b), we can validate the effectiveness of the stochastic Domain Guided Dropout when fine-tuning the CNN model. This is because we utilize the domain information to regularize the network better, which keeps the CNN in the right track when training data is not enough.

\begin{figure}[t]
\begin{center}
\includegraphics[width=0.8\linewidth]{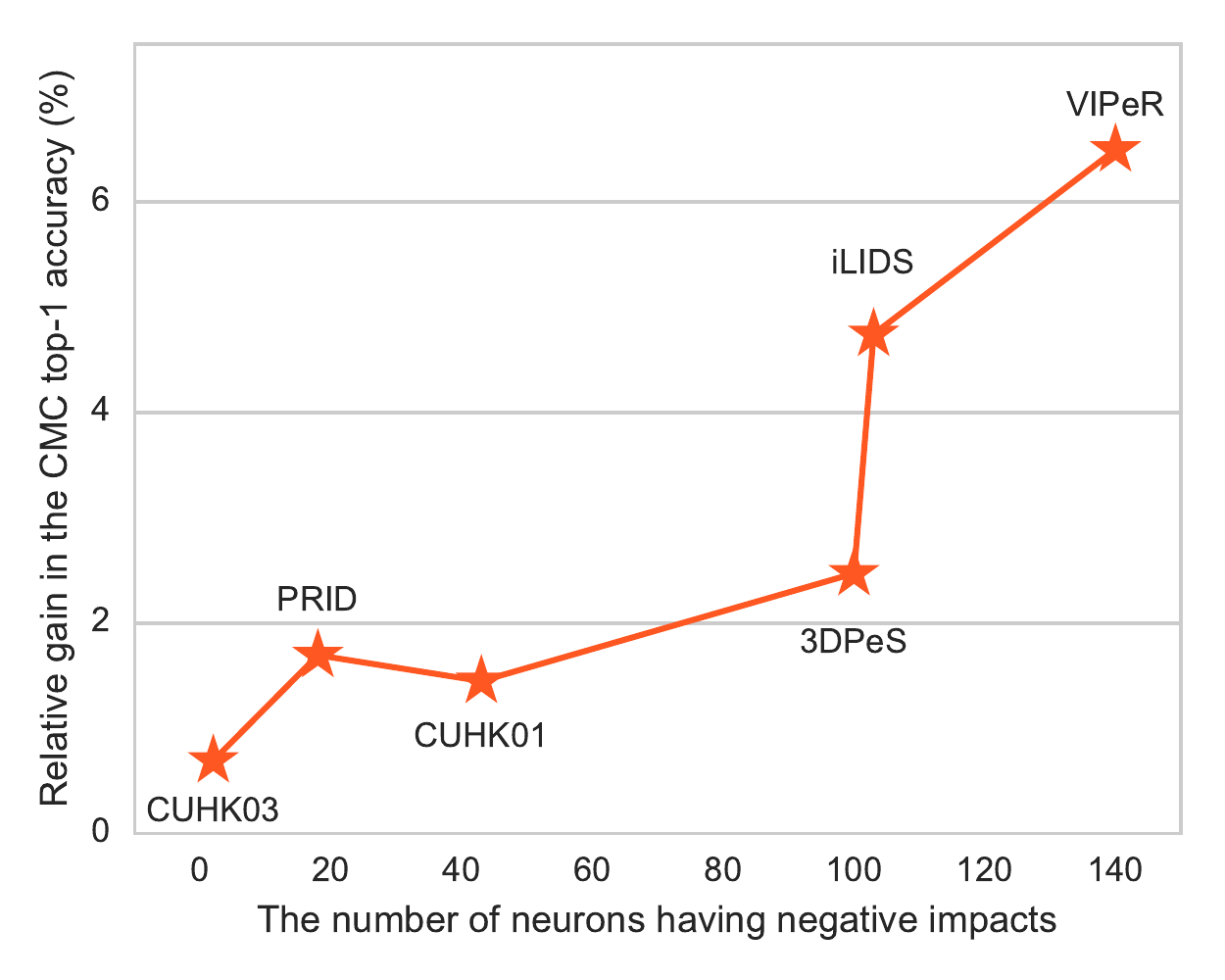}
\end{center}
\caption{Relative performance gain with respect to the number of neurons having negative impact scores on specific domain in the deterministic Guided Dropout scheme}
\label{fig:gain_wrt_neg_neurons}
\end{figure}

We further investigate how does the deterministic Domain Guided Dropout change the network behavior by evaluating the relative performance gain on each domain with respect to the number of neurons having negative impact scores on that domain. As shown in Figure~\ref{fig:gain_wrt_neg_neurons}, smaller datasets tend to have more useless neurons to be dropped out, meanwhile the performance would be increased more significantly. This again indicates that we should not treat all the domains equally when using all their data, but rather regularize the CNN properly for each of them.

% subsection effectiveness_of_the_guided_dropout_module (end)

% subsection comparison_with_state_of_the_art_results (end)

% section experiments (end)

\section{Conclusion} % (fold)
\label{sec:conclusion}

In this paper, we raise the question of learning generic and robust CNN feature representations from multiple domains. An effective pipeline is presented, and a Domain Guided Dropout algorithm is proposed to improve the feature learning process. We conduct extensive experiments on multiple person re-identification datasets to validate our method and investigate the internal mechanisms in details. Moreover, our results outperform state-of-the-art ones by large margin on most of the datasets, which demonstrates the effectiveness of the proposed method.

% section conclusion (end)

\section*{Acknowledgements}
\label{sec:acknowledgements}
This work is partially supported by SenseTime Group Limited, the General Research Fund sponsored by the Research Grants Council of Hong Kong (Project Nos. CUHK14206114, CUHK14205615, CUHK417011, CUHK14207814, CUHK14203015), and National Natural Science Foundation of China (NSFC, NO.61371192).

{\small
\bibliographystyle{ieee}
\bibliography{0110}
}

\end{document}